\documentclass[letterpaper, 10 pt, conference]{ieeeconf}  

\IEEEoverridecommandlockouts                              

\usepackage[noadjust]{cite}
\usepackage{amsmath,amssymb,amsfonts}
\usepackage{algorithmic}
\usepackage{graphicx}
\usepackage{textcomp}
\usepackage{xcolor}
\usepackage{caption}
\usepackage{multirow}
\usepackage{footnote}
\usepackage{hyperref}
\usepackage{array}
\usepackage{amssymb}
\usepackage{todonotes}
\usepackage[pscoord]{eso-pic}
\usepackage{flushend}

\title{\LARGE \bf
Towards Infield Navigation: leveraging simulated data for crop row detection
}

\author{Rajitha de Silva$^{1}$, Grzegorz Cielniak$^{2}$ and Junfeng Gao$^{3}$
\thanks{This work was supported by Lincoln Agri-Robotics as part of the Expanding Excellence in England (E3) Programme.}
\thanks{$^{1}$Rajitha de Silva, $^{2}$Grzegorz Cielniak and $^{3}$Junfeng Gao are with Lincoln Agri-Robotics Centre, Lincoln Institute for Agri-Food Technology, University of Lincoln, UK
        {\tt\small $^{1}$rajitha@ieee.org, $^{2}$gcielniak@lincoln.ac.uk, $^{3}$jugao@lincoln.ac.uk}}%
}

\begin{document}

\maketitle
\thispagestyle{empty}
\pagestyle{empty}

\begin{abstract}
Agricultural datasets for crop row detection are often bound by their limited number of images. This restricts the researchers from developing deep learning based models for precision agricultural tasks involving crop row detection. We suggest the utilization of small real-world datasets along with additional data generated by simulations to yield similar crop row detection performance as that of a model trained with a large real world dataset. Our method could reach the performance of a deep learning based crop row detection model trained with real-world data by using 60\% less labelled real-world data. Our model performed well against field variations such as shadows, sunlight and grow stages. We introduce an automated pipeline to generate labelled images for crop row detection in simulation domain. An extensive comparison is done to analyze the contribution of simulated data towards reaching robust crop row detection in various real-world field scenarios.
\end{abstract}

\section{Introduction}
Repetitive agricultural tasks such as weed control requires the robots to navigate through crop fields while accurately identifying the crop rows. Computer vision algorithms have been identified as one of the key areas that need to be improved to promote these autonomous agricultural systems  \cite{oliveira2021advances}. Crop detection is a key element in developing such autonomous agricultural systems.

Crop row detection has been a popular research question in classical computer vision. Researchers have developed colour based segmentation of crops with various methods with varying levels of success \cite{romeo2012crop},\cite{guerrero2013automatic}. Individual work has been done to pursue various challenges in implementing a vision based navigation algorithms in crop rows. Researchers have attempted to address the effects of weed density, growth stages, shadows and discontinuities in crop row detection \cite{ji2011crop}, \cite{fue2020evaluation}. While these attempts indicate successful solutions to each of those problems separately, the need of a generic algorithm arises in practical implementations in such systems.  A farmland may contain regions with varying levels of weed, various growth stages due to the soil conditions and different orientations of land. External environmental factors such as illuminations and shadows may also affect the performance of autonomous systems with the time of the day. Development of a generic crop row detection method under varying conditions is critical to achieve accurate autonomy in crop fields.

Recent work on crop row detection with deep learning based methods has been able to overcome the major challenges in implementing a real world vision based navigation system \cite{pang2020improved}, \cite{bah2019crownet}. Lacking of publicly available datasets limits the potential to develop such navigation methods the despite the proven effectiveness of those deep learning models. The existing datasets such as Crop Row Benchmark Dataset (CRBD) \cite{vidovic2016crop} lacks sufficient images to train deep learning models. The tedious nature of annotating such large datasets could be a reason for this.

In this paper, we present a method of using fewer annotated real world data in conjunction with simulated data to train a deep leaning model for crop row detection. The model was trained with different ratios of real and simulation data to evaluate the accuracy of this method. To this end, we have created a real world dataset representing the practical challenges encountered by a crop row detection system such as weed, shadows, discontinuities and grow stages. The annotated real world dataset with various field scenarios described in this paper is publicly available in order to further promote the development and optimisation of deep learning based row detection approaches for a reliable autonomous navigation system in fields. A simulation pipeline was developed to generate simulated training data with autonomous annotation, eliminating the need of manual annotation.

\section{Related Work}
Contour detection followed by Hough transform is a classic computer vision approach for crop row detection \cite{bonadies2019overview}. Hough transform based algorithms are popular among the researchers due to its accuracy of crop row detection \cite{winterhalter2018crop}, \cite{ji2011crop} and \cite{GAO201843}. However, these methods require fine tuning of threshold values to suit the varying ambient light conditions and growth stages. Thus, these classic approaches are not be suitable for real world vision-based navigation with varying environmental conditions. 

Ahmadi et al. \cite{ahmadi2020visual} has used excess green index  \cite{woebbecke1995color} based algorithm to detect crop rows. Living tissue indicator and vegetation index has also been used to develop crop row detection algorithms \cite{bakker2008vision}, \cite{montalvo2012automatic}. These color based indicators are used to isolate pixels corresponding to plants in an image. However, these algorithms find it challenging to perform well in the presence of weeds due to their inability to separate crops and weeds. These methods are also incapable of extracting crop rows where crop canopy closes hiding background pixels corresponding to soil.

Recent developments in machine learning have enabled to perform crop row detection using neural networks, overcoming some of the barriers in classical computer vision. Lane detection in autonomous driving on roads are proven the capability of deep learning to be used in real life autonomous navigation \cite{zhao2020deep}. Adhikari et al. \cite{adhikari2020deep} has developed a enhanced skip network based autonomous navigation system for paddy fields. Their system is robust against shadows, field of view, row spacing and growth stage of the crop. However their system is implemented under the assumption that crop rows are clearly distinguishable in an image. Mean Pixel Deviation metric is used to evaluate the model. Their dataset is only limited to 350 images and it does not cover all the different scenarios that can occur in a field.

Bah et al. \cite{bah2019crownet} has used a fully convolutional network architecture that combines SegNet \cite{badrinarayanan2017segnet} and HoughCNet which is a Hough transform on a skeletonized binary image followed by a convolutional neural network (CNN). Their approach tries to eliminate the effects of weeds and discontinuities on crop row detection with multiple convolutional network stages. This complex architecture has been able to provide accurate crop row detection performance on images captured by unmanned aerial vehicles (UAV). In most of the deep learning based crop row detection approaches, mean Intersection over union (IoU) is considered as the key performance metric.

Cerrato et al. \cite{cerrato2021deep} has used a set of realistic augmentations to generate simulated dataset in an orchard setting. They have generated a large dataset with sufficient augmentations to train a deep learning model to predict crops in a simulated environment. They are using a separate real-world dataset to make predictions in real world scenario. Our work attempts to bridge the gap between these two domains by using a common dataset with limited real-world data.

The existing work indicates the progress made in crop row detection over time and the advantages of deep learning based systems over classical computer vision approaches. The prior work on deep learning based crop row detection only analyze their model performance under one varying field parameter or without any variations, leading to the uncertainty of model deployment in field environments. The smaller datasets used in training such models indicate the absence of sufficient variations in those datasets to generalize over varying field parameters. Inflating the dataset with realistic augmentations in the simulation domain will provide the deep learning model with the ability to detect crop rows in varying field conditions.

\section{Dataset}

\begin{figure}[t]
\centering
\captionsetup{justification=centering}
\includegraphics[scale=0.22]{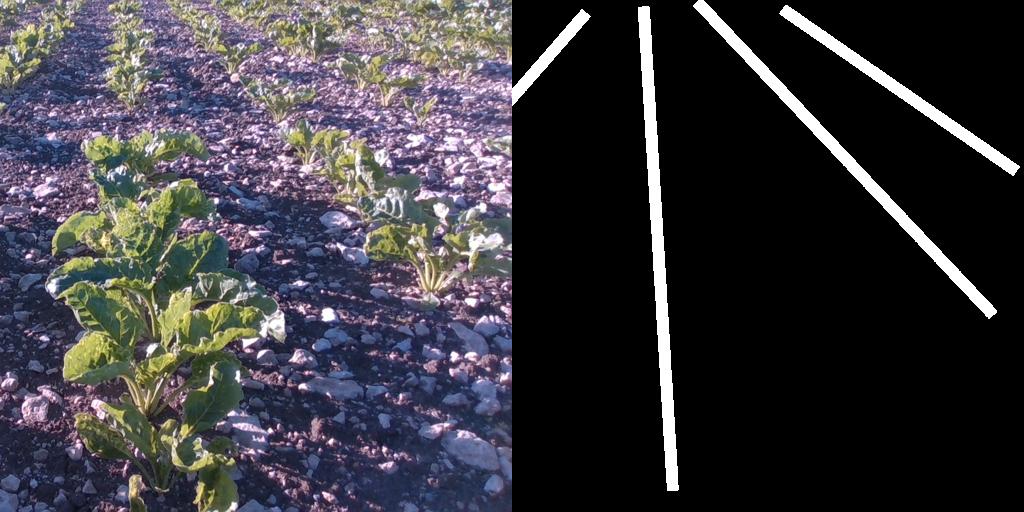}
\caption{Sample image and respective ground truth label mask}
\label{fig:lbl}
\end{figure}

The Crop Row Detection Lincoln Dataset (CRDLD) was created to gather a comprehensive real world training set for deep learning model which includes multiple possible under varying field conditions. The dataset includes variations in weed density, shadows, sunlight, terrain elevation, growth stages, shape of the crop row and discontinuities in crop rows. CRDLD consists of 2000 images augmented from 500 base images. Each base image is cropped in different orientations to generate four augmented images. The dataset was created from data recordings obtained in 3 days within a span of two weeks under varying weather conditions and different times of the day in a sugar beet field. The sugar beet plants had 4-10 unfolded leaves throughout the duration of data capturing. The 2000 image dataset was  split in half for training and testing. A larger test set is required since the model was evaluated in multiple categories of data. An example image and its corresponding ground truth image are shown in Figure \ref{fig:lbl}. The ground truth labels were annotated by identifying the start and end points of each crop row in the image. The width of each line in crop row annotation was 6 pixels. The expected outcome of our system is to predict the line formed by the arrangement of plants in an input image rather than segmenting individual plants. The crop row discontinuities due to missing plants are ignored and still assumed as a continuous crop line in the ground truth mask. 

\begin{figure}[t]
\centering
\captionsetup{justification=centering}
\includegraphics[scale=0.25]{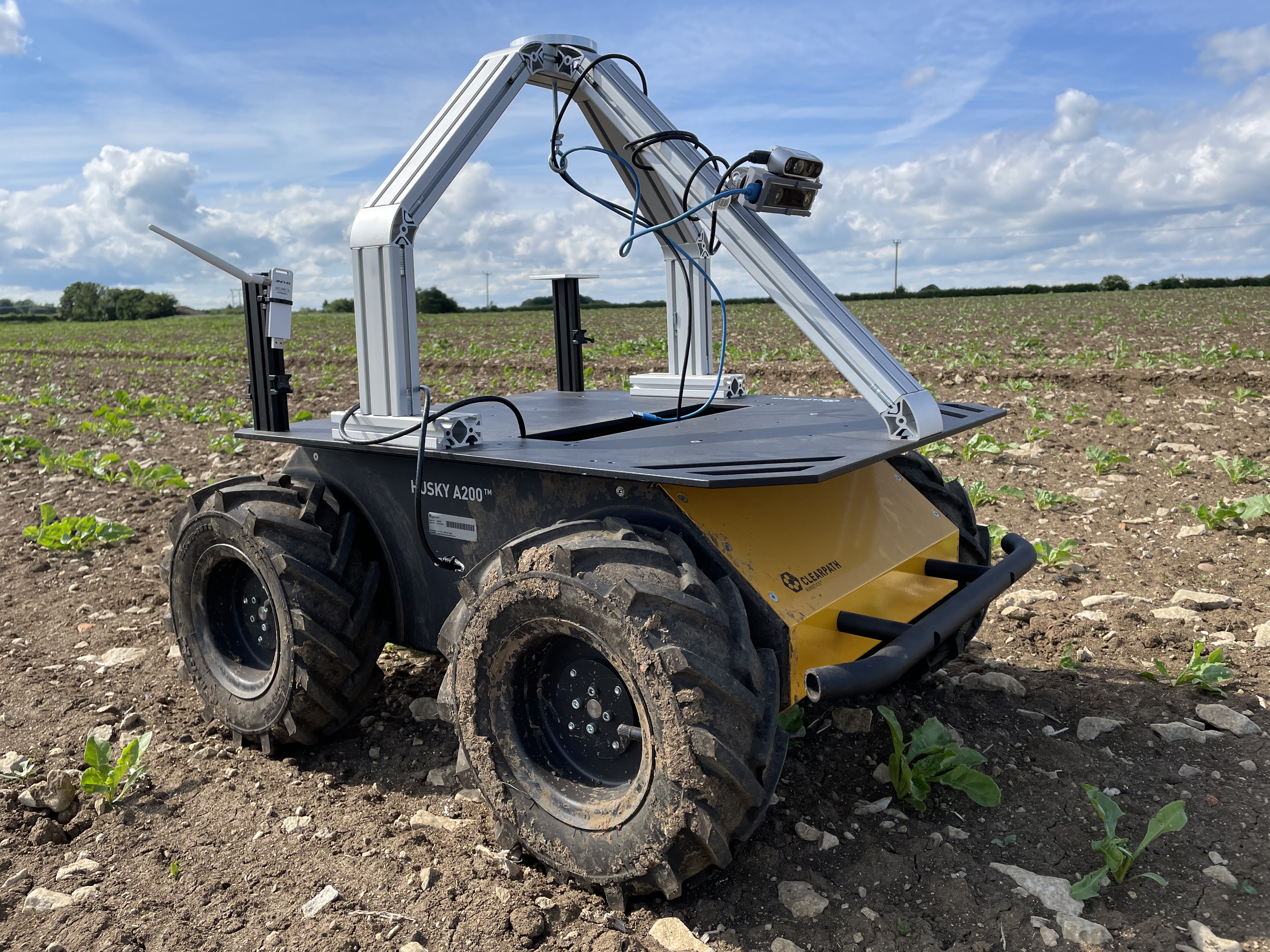}
\caption{Husky Robot with Realsense Cameras}
\label{fig:hus}
\end{figure}

The data collection was conducted using a Husky robot equipped with Intel RealSense D435i camera as shown in Figure \ref{fig:hus}. RGB, infra red (IR) and depth images were captured with D435i camera. Only RGB images from D435i camera is used to generate the dataset in this paper.

\subsection{Data Categories}
\begin{figure*}[t]
\centering
\captionsetup{justification=centering}
\includegraphics[width=6in]{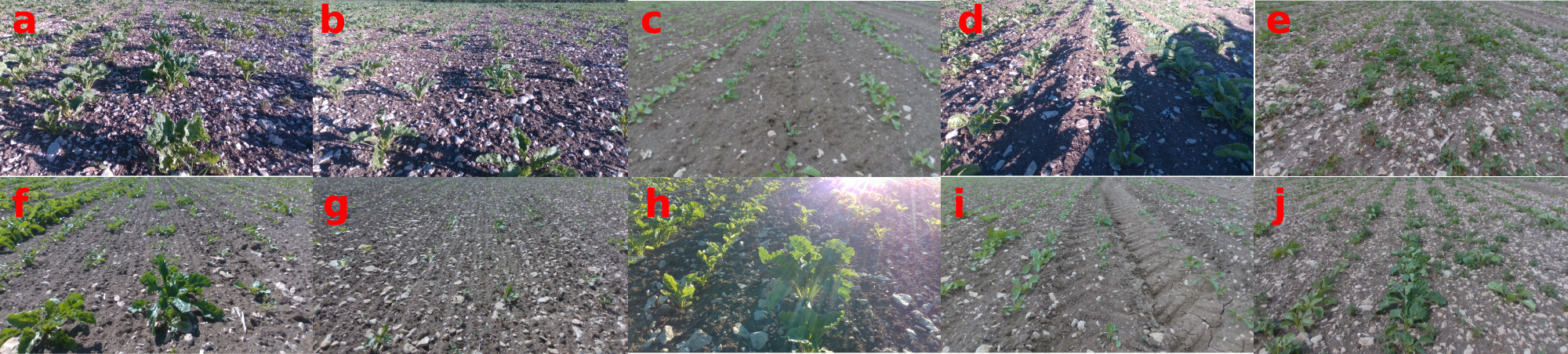}
\caption{Sample images from data categories in real-world dataset}
\label{fig:cat}
\end{figure*}

The RGB images in the dataset were classified into 10 categories depending on possible variations which could be expected in an open field farm due to varying weather, growth stages and time of the day. Each category contains 100 images along with the respective ground truth images\footnote{The dataset can be accessed with the following link: \textbf{\url{https://github.com/JunfengGaolab/CropRowDetection}}.}. The breakdown of categories is explained in Table \ref{tab:cat}. Sample images from each category are shown in Figure \ref{fig:cat}. 

\begin{table}[t]
\centering
\caption{Data Categories}
\begin{center}
\begin{tabular}{|p{0.04\linewidth} | p{0.2\linewidth} | p{0.55\linewidth}|}
\hline
\textbf{ID} & \textbf{Name} & \textbf{Description}\\
\hline
\multirow{2}{0.4em}{a} & \multirow{2}{0.25em}{Horizontal Shadow} & Shadow falls perpendicular to the direction of the crop row \\ 
\hline
\multirow{3}{0.4em}{b} & \multirow{3}{0.25em}{Slope/ Curve} & Images captured while the crop row is not in a flat farmland or where crop rows are not straight lines \\ 
\hline
\multirow{2}{0.4em}{c} & \multirow{2}{0.25em}{Discontinuities} & Missing plants in the crop row which leads to discontinuities in crop row \\ 
\hline
\multirow{2}{0.4em}{d} & \multirow{2}{0.25em}{Front Shadow} & Shadow of the robot falling on the image captured by the camera \\ 
\hline
e & Dense Weed & Weed grown densely among the crop rows \\ 
\hline
\multirow{2}{0.4em}{f} & \multirow{2}{0.25em}{Large Crops} & Presence of one or many largely grown crops within the crop row \\ 
\hline
g & Small Crops & Crop rows at early growth stages \\ 
\hline
\multirow{2}{0.4em}{h} & \multirow{2}{0.25em}{Sunlight} & Sunlight falling on the camera causing lens flares and similar distortions \\ 
\hline
\multirow{2}{0.4em}{i} & \multirow{2}{0.25em}{Tyre Tracks} & Tyre tracks from tramlines running through the field \\ 
\hline
\multirow{2}{0.4em}{j} & \multirow{2}{0.25em}{Sparse Weed} & Sparsely grown weed scattered between the crop rows \\ 
\hline
\end{tabular}
\label{tab:cat}
\end{center}
\end{table}

These 10 categories of data could be identified as general challenges that a robot will have to overcome to perform reliable vision based navigation in an outdoor field. There have been many attempts to solve these challenges in individual dimensions. Montalvo et al. \cite{montalvo2012automatic} has developed a system to detect crop rows in maize fields with high weed pressure, hence trying to solve the crop row detection problem in category "e" of the dataset. Sivakumar et al. \cite{sivakumar2021learned} has developed an under canopy navigation robot with a vision based system attempting to address the major variation of appearance between early and late growth stages. Their approach is an attempt to solve the crop row detection problem in the categories "f" and "g" of the dataset. A compendious solution for the crop row detection problem is yet to be explored, despite the availability of different solutions to provide good performance in individual contexts. To this end, availability of highly diverse training dataset is vital to achieve a vision based navigation implementation that can overcome the challenges posed in a real world environment. 

\section{Methodology}

The crop row detection algorithm is based on U-Net \cite{ronneberger2015u} architecture. U-Net has been one of the most popular image segmentation algorithm known for its ability to be trained with lesser amount of data and faster predictions. U-Net, being primarily intended for semantic segmentation, has been used to identify pixels belongs to a plant in a given image in agricultural applications \cite{fawakherji2019crop}, \cite{doha2021deep}. 

The U-Net was initially trained with the real world dataset of 750 images. The resulting model was tested on the test dataset containing 250 images composed of all the 10 categories listed in Table \ref{tab:cat}. This evaluation provides the insight into the ability of deep learning based crop row detection methods in handling external variables present in a real world crop field.

\subsection{Simulated Data Generation}
\begin{figure}[t]
\centering
\captionsetup{justification=centering}
\includegraphics[scale=0.15]{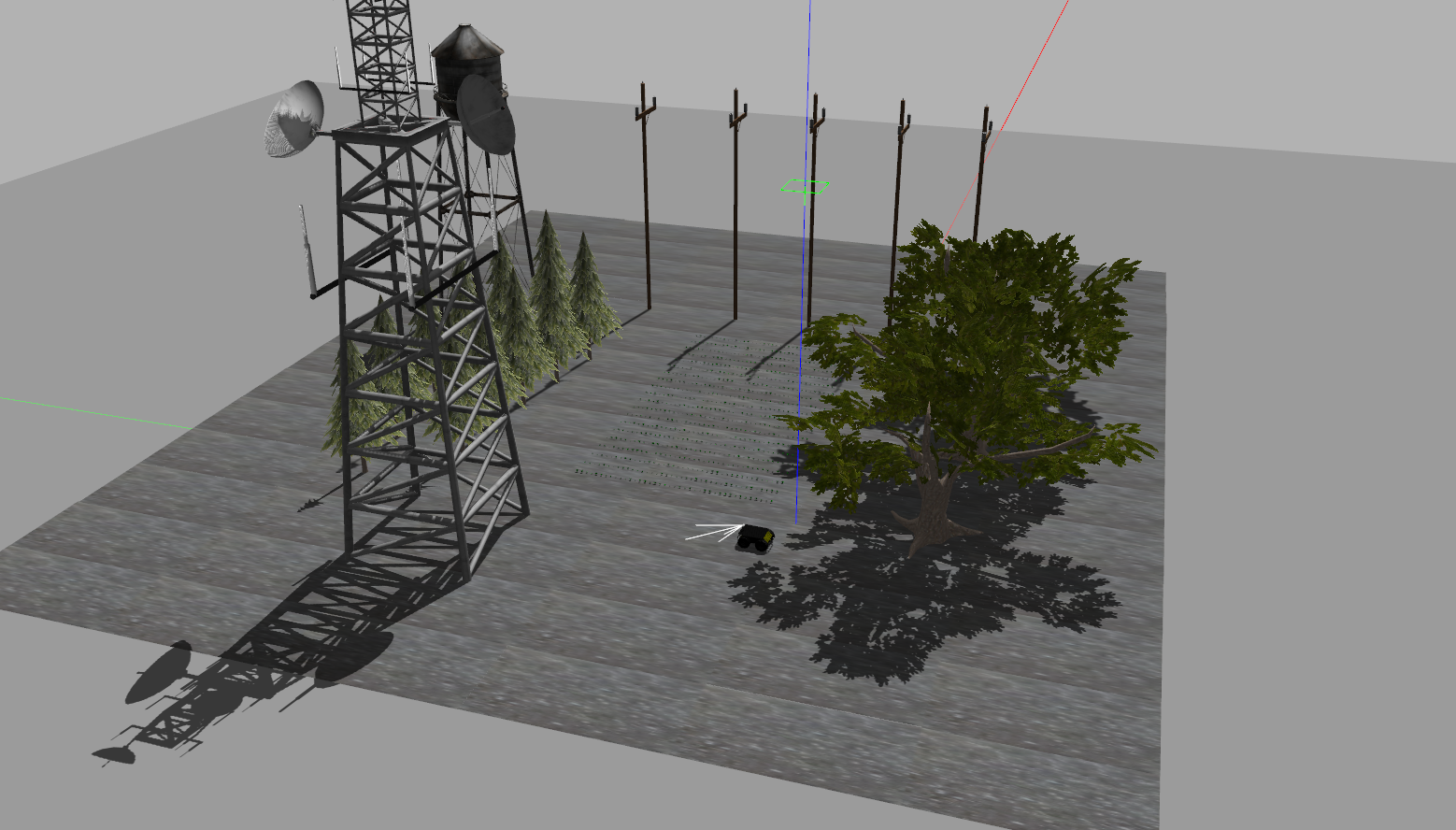}
\caption{Sugar Beet Field Simulation with Husky Robot}
\label{fig:smn}
\end{figure}

\begin{figure}[t]
\centering
\captionsetup{justification=centering}
\includegraphics[scale=0.12]{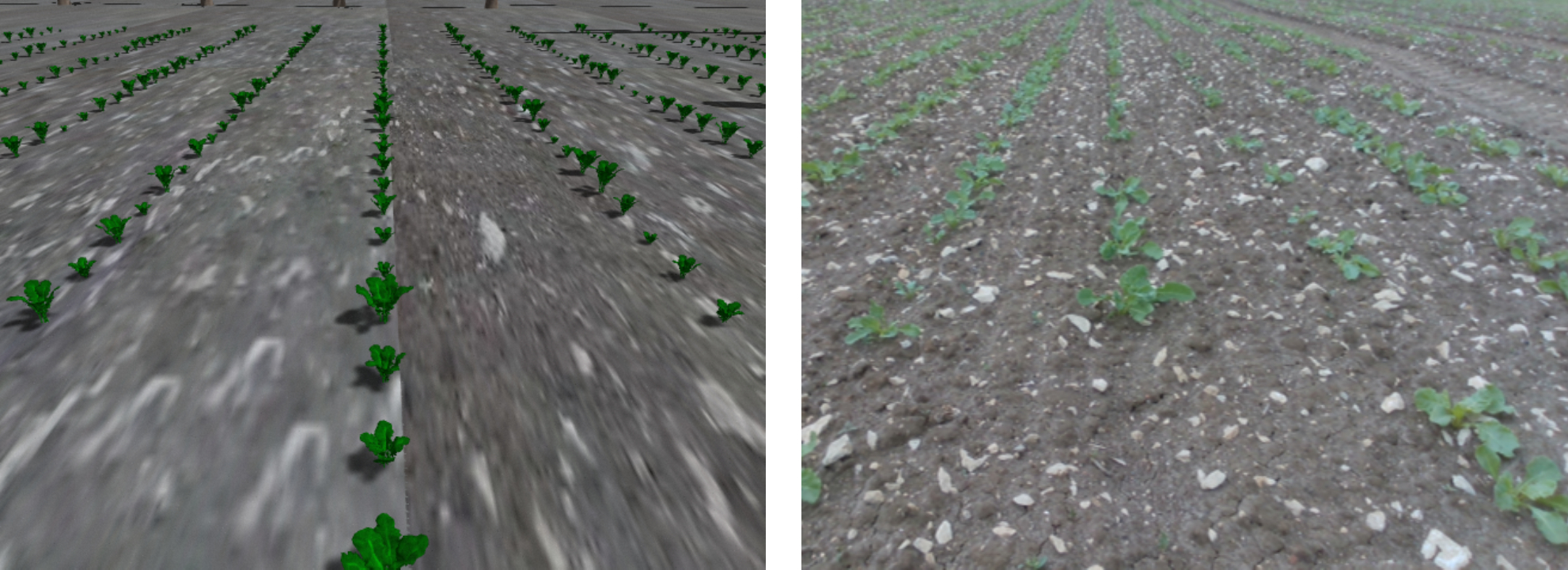}
\caption{Sample Image from Simulated Dataset (Left) and Sample Image from Real World Dataset (Right)}
\label{fig:svr}
\end{figure}

A crop row simulation was setup in Gazebo simulator as shown in Figure \ref{fig:smn}. The size and the z-axis rotation of the sugar beet plant is randomized to emulate the random nature of plants in a real field using "Randomized Transform" tool in Blender software. A summary of simulation parameters is listed in Table \ref{tab:smp}. Soil texture in the images of real world dataset was extracted to be used as the ground plane texture in the simulation. Leaves of the sugar beet plant 3D model in simulation was mapped with a realistic leaf texture to match the real sugar beet plant appearance as seen in the left image of Figure \ref{fig:svr}. The texture mapping leads the simulation images to be more realistic represented in Figure \ref{fig:svr}. The robot is driven through a sequence of pre-defined way-points in the simulation where the robot captures an image at each way-point.

\begin{table}[t]
\centering
\caption{Simulation Parameter for Sugar Beet Field}
\begin{center}
\begin{tabular}{|p{0.25\linewidth} | p{0.12\linewidth} | p{0.14\linewidth}| p{0.17\linewidth}|}
\hline
\textbf{Property} & \textbf{Value} & \textbf{Variance}\\
\hline
{Row Spacing} & 60cm & 0 \\ 
\hline
{Seed Spacing} & 16cm & 0 \\ 
\hline
{Plant Height} & 6cm & +3cm \\ 
\hline
{Plant Orientation} & 0\textdegree & $\pm$ 145\textdegree \\ 
\hline
Row Length & 6m & 0 \\ 
\hline
{Row Count} & 20 & 0 \\ 
\hline
\end{tabular}
\label{tab:smp}
\end{center}
\end{table}

\begin{figure}[t]
\centering
\captionsetup{justification=centering}
\includegraphics[scale=0.1]{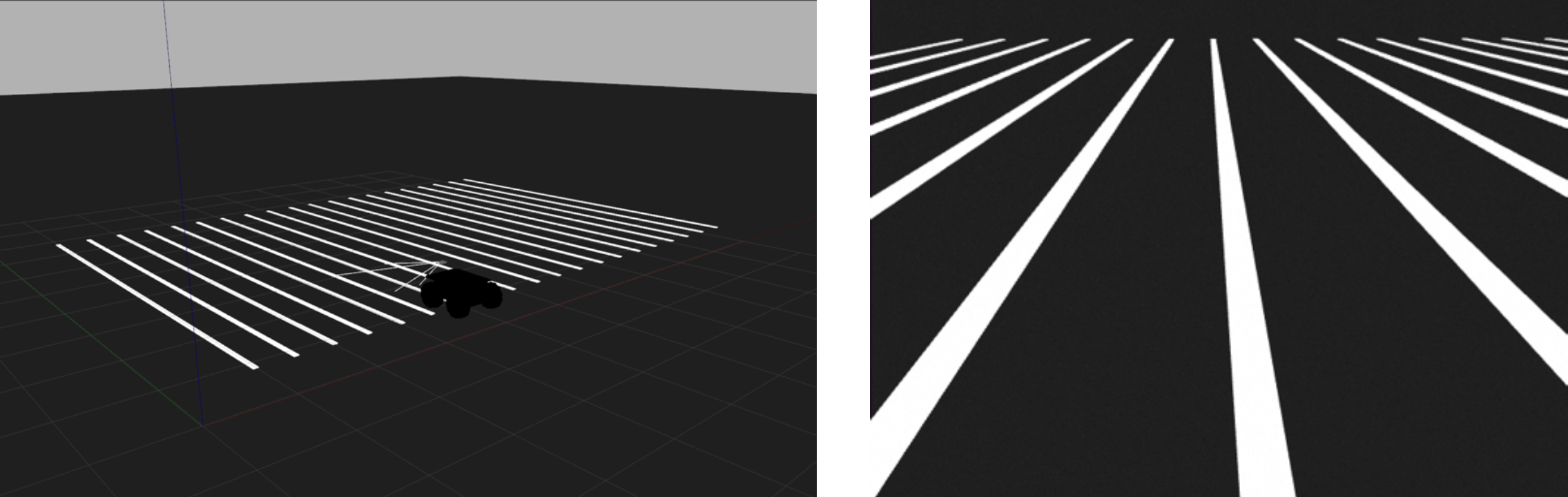}
\caption{Ground Truth Simulation (Left) and a Sample Annotation from Ground Truth Simulation Dataset (Right)}
\label{fig:gts}
\end{figure}

The ground truth labels corresponding to the images obtained in the simulation are autonomously generated with the aid of a "Ground Truth Simulation (GTS)". The GTS is a mirror simulation environment of the original simulation where crop rows are replaced with rectangular stripes emitting uniform white colour light and all the other objects in the original simulation are removed, including the global light source. The robot is driven in the GTS through the same way-points and stopped at each point to capture images in the original simulation. The GTS and a sample image captured in GTS is shown in Figure \ref{fig:gts}. 

This data generation pipeline with dual simulation environment is enabling faster access to large annotated datasets, since a user could control the number of plants, crop rows and all the other environment variations in a simulation environment. 

\subsection{Training with Real World Data}
The U-Net was trained with a dataset of 750 images which comprised of 75 images from each category listed in Table \ref{tab:cat}. Image resolution of both input and ground truth images was $512\times512$ pixels. The model was trained with binary cross entropy (BCE) loss with the Adam optimizer at a learning rate of $1\times10^{-4}$. The model was also trained separately with focal loss with $\gamma$ = 2. The predicted crop row masks were more sharper and narrower with the BCE loss. Therefore the BCE loss was selected for the work described in following sections.

\begin{figure}[t]
\centering
\captionsetup{justification=centering}
\includegraphics[scale=0.09]{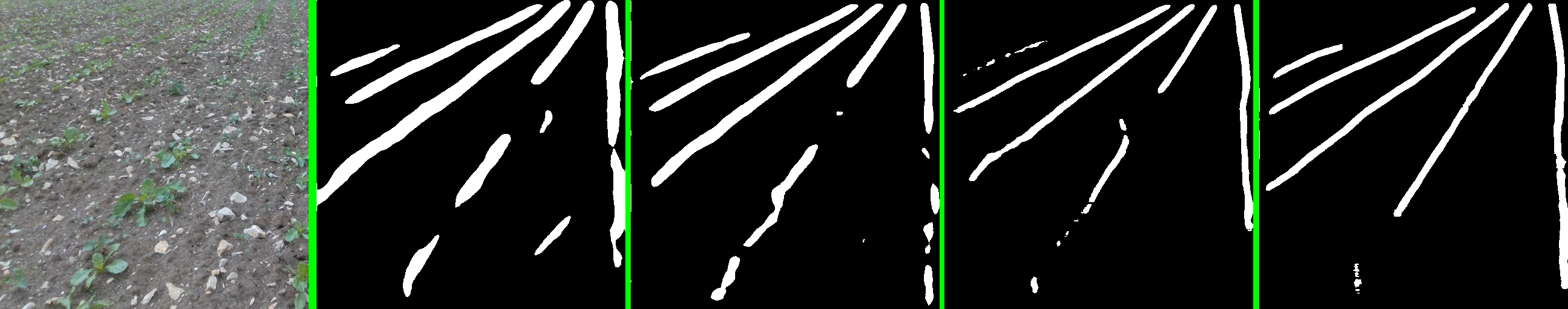}
\caption{Improvement of prediction at 5, 10, 20 and 40 epochs respectively (Left to Right)}
\label{fig:timp}
\end{figure}

As shown in Figure \ref{fig:timp}, U-Net has started to recognise the crop rows with only 5 epochs of training. However, this early stage model is only able to predict the line when there are no discontinuities in the crop row. It does not detect the crop row in regions where discontinuities are seen in crop rows. It is also noticeable that the predicted crop row is relatively wider than the ground truth mask. The model was able to eliminate false positives when training up to 10 epochs while it became more narrow and straight when trained up to 20 epochs. The network was trained at 40 epochs where model could predict the entire crop rows by overcoming the aforementioned challenges such as crop discontinuities. The model accuracy saturated after 40 epochs.

\subsection{Training with Simulated Data}

Another instance of the U-Net was trained only using simulated images generated by the simulation pipeline described in Section IV.A. This simulation data based model could predict the general crop row structure in real world images even without seeing a single real world image during training. However, these predictions were not sharp and clear as in the model trained with real world data. Since this model was able to predict the general crop row structure in real world data, adding a small number of real world data in conjunction with the simulation based dataset would yield more accurate predictions of the crop row in real world data. This hypothesis is tested and evaluated to understand the real world data prediction performance vs. number of annotated real world images trade-off in training the U-Net model for crop row detection. 

An initial experiment was conducted to infer the possibility of training U-Net with simulated crop row images to predict crop rows in real world environment. Successive instances of U-Net were trained with varying real world and simulated image combinations as listed in Table \ref{tab:data}. All the models were trained with same hyperparameters to observe the effect of training data combinations on model performance.

\begin{table}[ht]
\centering
\caption{Training Data Variations for Training U-Net}
\begin{center}
\begin{tabular}{|p{0.12\linewidth} | p{0.19\linewidth} | p{0.19\linewidth}| p{0.12\linewidth}|}
\hline
\textbf{Model ID} & \textbf{Simulated Images} & \textbf{Real-World Images} & \textbf{IoU (\%)}\\
\hline
A1 & 500 & 0 & 6.93 \\ 
\hline
A2 & 500 & 50 & 13.81 \\ 
\hline
A3 & 500 & 100 & 16.75 \\ 
\hline
A4 & 500 & 150 & 16.10 \\ 
\hline
A5 & 500 & 200 & 18.83 \\ 
\hline
A6 & 500 & 250 & 19.24 \\ 
\hline
B1 & 1000 & 0 & 7.22 \\ 
\hline
B2 & 1000 & 100 & 15.75 \\ 
\hline
B3 & 1000 & 200 & 17.31 \\ 
\hline
B4 & 1000 & 300 & 19.71 \\ 
\hline
B5 & 1000 & 400 & 20.04 \\ 
\hline
B6 & 1000 & 500 & 21.28 \\ 
\hline
R & 0 & 750 & 22.50 \\ 
\hline
\end{tabular}
\label{tab:data}
\end{center}
\end{table}

\section{Experimental Evaluation}

The U-Net was trained with different data combinations listed in Table \ref{tab:data} and the IoU (Intersection over Union) for validation dataset was monitored for each model. The models were trained under 3 categories A, B and R. Category A has only 500 simulated images in the training set while category B has 1000 simulated images. The Category R model was only trained on real world images. The model R performance is considered as the benchmark performance to evaluate the models in categories A and B.

\subsection{IoU Analysis}

Accuracy and IoU metrics are commonly used to evaluate the performance of a semantic segmentation model. The percentage in which the model predicts each pixel in an image similar to the corresponding ground truth image pixel is the accuracy in a binary segmentation problem. In crop row detection problem, a high accuracy score does not represent a good performance of a model due to large number of background pixels in comparison to crop row pixels. The calculation of IoU is given in Equation \ref{eq:1} where TP is true positive pixel count, FP is false positive pixel count and FN is false negative pixel count. IoU is the most accepted metric for semantic segmentation model evaluation due to its reflection on detecting the desired elements rather than the entire image. This is especially true in our case as the  pixels belonging the two classes (crop row, background) are quite imbalanced. Therefore, IoU is used to evaluate the performance of all the models. 

\begin{equation} \label{eq:1}
 IoU = \frac{TP}{TP+FP+FN}
\end{equation}

\begin{figure}[t]
\centering
\captionsetup{justification=centering}
\includegraphics[scale=0.185]{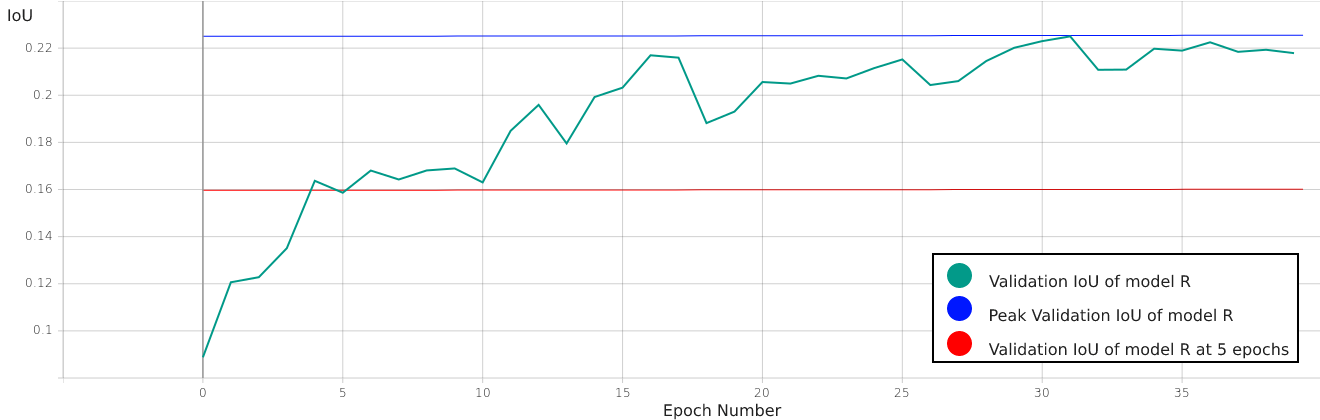}
\caption{Validation IoU variation during training for model R }
\label{fig:rvaliou}
\end{figure}

The U-Net model R gained the ability to predict crop rows at 5 epochs of training as indicated in Figure \ref{fig:timp}. Therefore, the baseline IoU for crop row detection is identified as 16\% according to the validation IoU curve for model R is given in Figure \ref{fig:rvaliou}. The peak IoU value for model R was recorded as 22.5\%. A model performance score $P_{m}$ was calculated to quantify the ability of U-Net models in categories A and B to predict crop rows. 

\begin{equation} \label{eq:2}
P_{m} = \left(\frac{\theta_{m}-\theta_{b}}{\theta_{p}-\theta_{b}}\right)\times100\%
\end{equation}

\begin{figure}[t]
\centering
\captionsetup{justification=centering}
\includegraphics[scale=0.3]{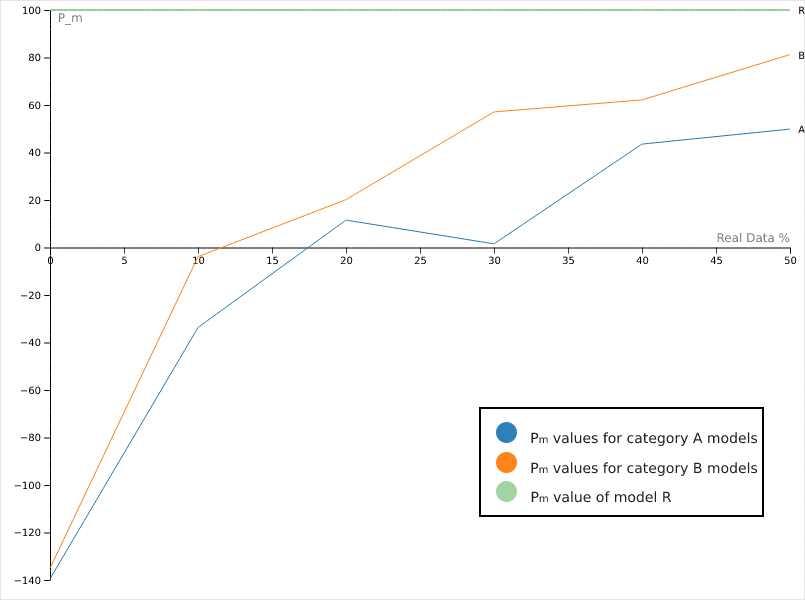}
\caption{$P_{m}$ vs. Real world image count as a percentage of simulated data in training set for model categories A and B}
\label{fig:pmc}
\end{figure}

$\theta_{m}$ is the IoU value of model $m$ in equaltion \ref{eq:2}. $\theta_{p}$(=0.225) and $\theta_{b}$(=0.160) are the peak and baseline IoU values of model R identified in Figure \ref{fig:rvaliou}. Figure \ref{fig:pmc} illustrates the variation of $P_{m}$ in model categories A and B with respect to relative percentage of real-world images presented in the training dataset. The relative percentage is the expression of real-world data count in the training dataset as a percentage of simulated data count. A minimum relative percentage of 20\% should represent in a training dataset for a model to be able to successfully predict crop rows. The trend of increasing $P_{m}$ indicates the obvious fact that a model will predict crop rows well when the number of real-world data count is increased. However, the experiment was stopped at 50\% relative percentage to preserve the key interest of this research: minimize the need of real-world data for crop row detection. Despite the lower $P_{m}$ values corresponding to lesser IoU, any model with a positive $P_{m}$ value could be considered as a successful crop row detection model to be used in a real life scenario. This could be further verified by the discussion Section $V.B$.

\subsection{Categorical Analysis}
The predictions from the models in categories A, B and R were observed to understand the response of each model towards the varying field conditions stated in Table \ref{tab:cat}. A summary of responses from the models in categories A, B and R towards varying field conditions is presented in Table \ref{tab:pcat}. The success of a model in each category is determined by the baseline IoU of 16\%. Each model is scored based on the number of data categories which the model achieves the succession criteria (16\%).

\begin{table*}[t]\centering
\caption{Positive $P_{m}$ Model Performance in Data Categories}
\begin{center}
\begin{tabular}{|p{0.18\linewidth} | p{0.05\linewidth} | p{0.05\linewidth}| p{0.05\linewidth}| p{0.05\linewidth} | p{0.05\linewidth}| p{0.05\linewidth}| p{0.05\linewidth} | p{0.05\linewidth}| p{0.05\linewidth}|}
\hline
\textbf{Category Name} & \textbf{R (\%)} & \textbf{A3 (\%)} & \textbf{B3 (\%)} & \textbf{A4 (\%)} & \textbf{B4 (\%)} & \textbf{A5 (\%)} & \textbf{B5 (\%)} & \textbf{A6 (\%)} & \textbf{B6 (\%)}\\
\hline
{Horizontal Shadow} & 15.88 & 12.19 & 14.68 & 12.09 & 16.04 & 15.13 & 15.54 & 14.19 & 15.52 \\ 
\hline
{Slope/ Curve} & 16.02 & 12.08 & 11.77 & 9.93 & 16.14 & 13.66 & 15.74 & 13.40 & 15.91 \\ 
\hline
{Discontinuities} & 25.44 & 22.05 & 22.01 & 21.23 & 23.94 & 24.11 & 23.93 & 24.47 & 25.20 \\ 
\hline
{Front Shadow} & 22.05 & 19.26 & 20.90 & 22.03 & 21.40 & 21.00 & 20.04 & 21.56 & 23.26 \\ 
\hline
Dense Weed & 17.69 & 8.83 & 6.69 & 3.81 & 12.13 & 7.82 & 12.89 & 8.32 & 16.15 \\ 
\hline
{Large Crops} & 29.47 & 19.75 & 21.60 & 19.15 & 24.65 & 22.98 & 26.00 & 24.87 & 27.50 \\ 
\hline
Small Crops & 30.49 & 24.15 & 25.06 & 24.28 & 26.15 & 26.63 & 27.41 & 26.73 & 28.66 \\ 
\hline
{Sunlight} & 22.29 & 19.95 & 19.43 & 20.73 & 20.85 & 22.42 & 21.41 & 22.19 & 21.95  \\ 
\hline
{Tyre Tracks} & 21.78 & 12.76 & 12.19 & 10.66 & 16.70 & 14.17 & 14.76 & 14.14 & 15.09 \\ 
\hline
{Sparse Weed} & 20.26 & 11.87 & 10.80 & 7.59 & 16.55 & 13.18 & 16.92 & 13.78 & 17.63 \\ 
\hline
\textbf{Model Score} & \textbf{9} & \textbf{5} & \textbf{5} & \textbf{5} & \textbf{9} & \textbf{5} & \textbf{6} & \textbf{5} & \textbf{7} \\ 
\hline
\end{tabular}
\label{tab:pcat}
\end{center}
\end{table*}

\begin{figure*}[ht]
\centering
\captionsetup{justification=centering}
\includegraphics[scale=0.25]{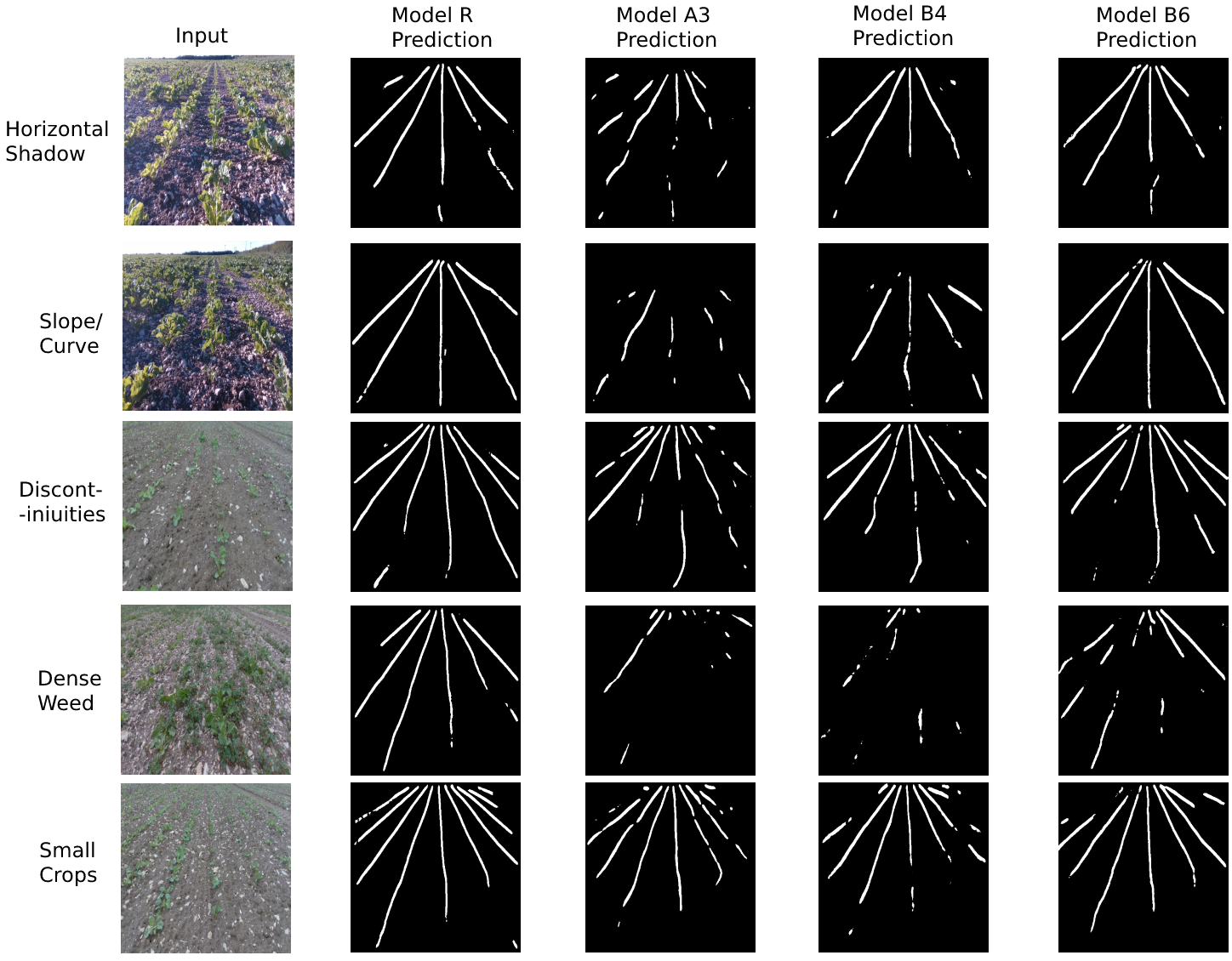}
\caption{ Prediction results from the models R, A3, B4 and B6}
\label{fig:dcc}
\end{figure*}

The models R and B4 tied for the high score successfully predicting crop rows in 9 out of 10 data categories. Model B6 scored 7 despite obtaining highest overall IoU (21.28\%) among both A and B model categories. The lowest recorded score was 5 indicating that all the models which had a positive $P_{m}$ score could predict crop rows successfully at least in 50\% of the field variation scenarios. The highest failure rate was recorded in "Dense Weed" data category and lowest failure rate was recorded in "small crops" data category. The simulated images had high resemblance to the "small crops" data category, hence the high IoU could be expected. The simulation had no images with dense weed presence to emulate the "Dense Weed" category data. Figure \ref{fig:dcc} presents a result comparison among the models R, A3, B4 and B6 in a few of interesting data categories. Model R being the benchmark model, models A3 and B6 had the lowest and highest overall IoU values. Model B4 was capable of predicting crop rows in most categories despite having a lesser overall IoU compared with B6. This comparison demonstrates the differences between each of these models highlighting their strengths and weaknesses. 

\section{Conclusion}
In this paper, we present a comprehensive real-world dataset for crop row detection including different field variations expected in a real field environment. We also introduce an automated labelling method for simulated crop row data generation in Gazebo simulator. The U-Net CNN could reach to a similar prediction performance of model R with only using 40\% of real-world training data with our method. The simulation data based models performed poorly in the presence of weed. This lack of performance could be justified by the absence of weed in the simulation. The simulation based models were robust against variations due to sunlight, shadows and grow stages. The experiments suggest that our approach could be used for accurate crop row detection without needing a large real-world dataset.

\bibliography{root}
\end{document}